\documentclass{article}

\usepackage{arxiv}

\usepackage[utf8]{inputenc} 
\usepackage[T1]{fontenc}    
\usepackage{hyperref}       
\usepackage{url}            
\usepackage{booktabs}       
\usepackage{amsfonts}       
\usepackage{nicefrac}       
\usepackage{microtype}      
\usepackage{lipsum}
\usepackage{graphicx}
\graphicspath{ {./images/} }

\usepackage{graphicx} 
\usepackage{multicol}
\usepackage{float}
\usepackage{arabtex}
\usepackage{utf8}
\setcode{utf8}
\usepackage[export]{adjustbox}
\usepackage{tabularx,rotating}
\usepackage{booktabs}
\usepackage{multirow}
\usepackage{caption}
\usepackage{subcaption}
\usepackage{comment}
\usepackage{rotating}
\usepackage{lscape}

\usepackage[dvipsnames]{xcolor}

\title{Content-Localization based Neural Machine Translation for Informal Dialectal Arabic: Spanish/French to Levantine/Gulf Arabic}

\author{
 Fatimah Alzamzami \\
  School of Electrical Engineering and Computer Science\\
  University of Ottawa\\
  Ottawa, Ontario, Canada \\
  \texttt{falza094@uottawa.ca} \\
   \And
 Abdulmotaleb El Saddik \\
  School of Electrical Engineering and Computer Science\\
  University of Ottawa\\
  Ottawa, Ontario, Canada \\
  Mohamed bin Zayed University of Artifcial Intelligence\\
  Abu Dhabi, UAE\\
  \texttt{elsaddik@uottawa.ca} \\
}

\begin{document}
\maketitle
\begin{abstract}
Resources in high-resource languages have not been efficiently exploited in low-resource languages to solve language-dependent research problems. Spanish and French are considered high resource languages in which an adequate level of data resources for informal online social behavior modeling, is observed. However, a machine translation system to access those data resources and transfer their context and tone to a low-resource language like dialectal Arabic, does not exist. In response, we propose a framework that localizes contents of high-resource languages to a low-resource language/dialects by utilizing AI power. To the best of our knowledge, we are the first work to provide a parallel translation dataset from/to informal Spanish and French to/from informal Arabic dialects. Using this, we aim to enrich the under-resource status of dialectal Arabic and fast-track the research of diverse online social behaviors within and across smart cities in different geo-regions. The experimental results have illustrated the capability of our proposed solution in exploiting the resources between high and low resource languages and dialects. Not only this, but it has also been proven that ignoring dialects within the same language could lead to misleading analysis of online social behavior.
\footnote{\scriptsize \textbf{Disclaimer:} This work uses terms, sentences, or language that are considered foul or offensive by some readers. Owing to the topic studied in this thesis, quoting toxic language is academically justified but I do not endorse the use of these contents of the quotes. Likewise, the quotes do not represent my opinions and I condemn online toxic language.}
\end{abstract}


\section{Introduction}
\label{intro}

With the ever-growing amount of user-generated multilingual content on social media, uncovering the valuable knowledge hidden within this information is crucial \cite{tracing-OSN,dst}. Examining such multilingual (i.e., including various dialects within the very same language) online social interactions is an essential tool for analyzing the social and cultural fabric of smart cities, allowing for informed planning and development strategies. Research trend, however, tends to solve a specific problem in a specific language or dialect, instead of utilizing the already existing resources to analyze online social behaviors (OSB), like sentiment and hate speech \cite{john2017sentiment, hate-hind, alruily2020sentiment, alzamzami2021monitoring}, in low-resource languages and dialects. This can be seen in the huge discrepancy of resources between languages, where very few have a high-resource status while many others are low-resource. Spanish and French are high-resource languages while Arabic is still considered a low-resource language \cite{sajjad2020arabench} despite the fact that Arabic is among the top used languages on social media following Spanish and French. This research trend encourages building resources for every problem in every language and dialect \cite{french-tweets-dataset, baly2019arsentd, dfsmd, haddad2019t, farha2020arabic, abdul2014sana, mulki2021let}
 and limits exploiting the already existing resources -especially in high-resource languages- to other different languages and even dialects to solve tasks like online social behavior analysis in low-resource languages. Moreover, the expensive cost (i.e. in terms of HW/SW requirements, time, efforts, cost, and human labor) \cite{alzamzami2020light} for building a resource for each language/dialect to solve each OSB task has definitely contributed to the under-resource status of existing multi-lingual-dialectal OSB systems. Researchers are obligated to address the shortcomings of this current practice and study reliable yet inexpensive solutions that minimizes the dependency of languages/dialects in modeling online social behaviors (e.g. sentiment and hate speech).

Machine translation is a medium to bridge the informal-communication gap between high and low resource languages on social media; however, word-to-word translation does not ensure transferring the context and tone of messages from a language/dialect to another language/dialect. Further, the unavailability of informal parallel datasets has contributed to the sub-optimal performance of machine translation in low-resource languages. This paper addresses this issue and proposes a content-localization based translation dataset (SF-ArLG) that provides parallel translations from/to informal Spanish and French to/from two informal Arabic dialects: Levantine and Gulf. This dataset is customized for informal communications on social media; it does not only consider the dialectal aspect of a language but also considers the social media culture for communications. We propose our SF-ArLG dataset as an attempt to minimize the dependency of languages/dialects in modeling online social behaviors in low-resource languages, through efficiently allowing the exploitation of existing resources especially in high-resource languages like Spanish and French. This is done through localizing the informal content of those existing resources into the low-resource informal dialectal Arabic without the need to construct and annotate new data resources from scratch. 
To the best of our knowledge, we are the first to construct an informal translation dataset from/to Spanish and French to/from Arabic Levantine and Gulf dialects. 
This paper is designed to answer the following questions: (1) Can we transfer the context and tone of social-informal messages between languages and dialects using AI power?, (2) Can we exploit existing OSB resources from a high-resource language/dialect to solve an online social behavior problem in a low-resource language/dialect using content-localization translation approach?. To the best of our knowledge, this is the first work that proposes a content-localization based machine translation framework to model online social behavior in low-resource languages/dialects.
We summarize the contributions of this work as follows:\\
(1) Design a content-localization based machine translation framework for online social behavior modeling in different language and dialects. \\
(2) Construct a parallel social-informal translation dataset between Spanish, French and dialectal Arabic.\\
(3) Develop neural machine translation models from/to French, Spanish to/from dialectal Arabic (i.e. Levantine and Gulf in this paper).\\
(4) Conduct comprehensive experiments to evaluate the performance of our proposed content-localize models on two classifications problems: sentiment classification and hate speech classification.

The rest of the paper is organized as follows. Section \ref{rw}
presents the related work. Our proposed method is presented in
Section \ref{method}. Section \ref{exp} explains
the experiment design and evaluation protocol followed in
this work whereas the results and analysis are discussed
in Section \ref{res}. Finally, in Section \ref{conc} we conclude our
proposed work and findings. Future directions are discussed in the limitation section.
\section{Related Work}
\label{rw}

Neural machine translation (NMT) has grown rapidly in the last ten years; it has shown a tremendous performance on high-resource language pairs, however, NMT is still inaccurate on low-resource languages due to the unavailability of large parallel datasets \cite{ranathunga2023neural}.
Even though the accuracy of translation from/to Arabic into/from French and Spanish is significantly lower than the accuracy for other language pairs such as English-French and English-Spanish  \cite{ranathunga2023neural, el2022readability}, there has been a notable progress in the development of machine translation from/to modern standard Arabic (MSA) into/from English language \cite{alqudsi2019hybrid, alqudsi2014arabic, zakraoui2020evaluation}. However, dialectal Arabic has not received the same level of attention as MSA in machine translation from different languages despite its dominant use over MSA on social media \cite{sajjad2020arabench}. Little effort has been noticed in dialectal Arabic machine translation from English language including Zbib etal \cite{zbib2012machine} (English-Levantine and English-Egyptian), MDC corpus \cite{bouamor2014multidialectal} (English-Levantine/North\-African/Egyptian), MADAR Corpus \cite{bouamor2018madar} (English-Gulf/Levantine/Egyptian-Nort\-African), QAraC \cite{elmahdy2014development} (English-Gulf), The Bible \footnote{\label{bible1} https://www.biblesociety.ma}, \footnote{\label{bible2} https://www.bible.com} (English-North\_African). To the best of our knowledge, there is no existing dataset that provides parallel translation between informal Spanish,  French, and dialectal Arabic. This paper proposes the first translation dataset from informal Spanish and French to informal Arabic dialects (i.e. Levantine and Gulf dialects in this paper).

\section{Method}
\label{method}
\begin{figure}[h]
\centering
\includegraphics[width=.426\textwidth]{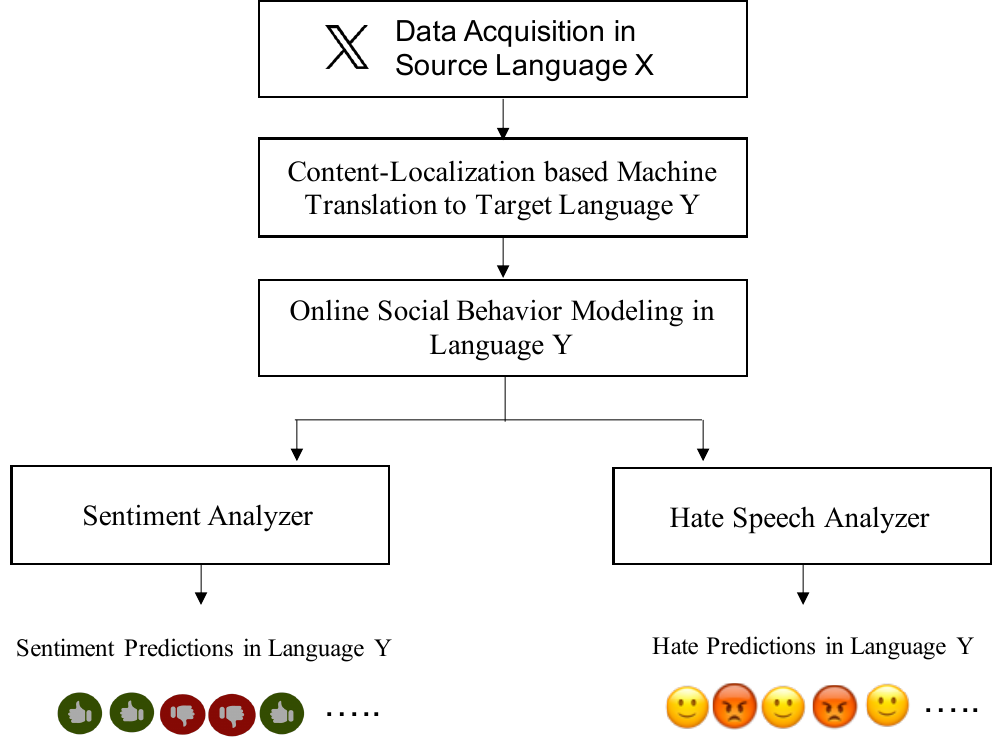}
\caption{Proposed framework.}
\label{fig:framework-general}
\end{figure}

Figure. \ref{fig:framework-general} presents the proposed framework for content-localization based neural machine translation approach for modeling online social behaviors (i.e. sentiment and hate speech as case studies in this work) in two low-resource Arabic dialects: Levantine and Gulf. 
The framework depicted in Figure. \ref{fig:framework-general} is proposed as an attempt to enrich the Arabic under-resource status so we can expand the spectrum of online social behavior analysis by minimizing the dependency of languages and dialects in modeling online social behaviors. This framework is designed in a way that reduces the cost of building resources (i.e. data and/or models) for every language and dialect to perform OSB-related tasks by allowing efficient exploitation of existing resources (i.e. data and/or models) between low-resource (like Arabic and its dialects) and high-resource languages (like French and Spanish).
To achieve this, we propose translating social media data, namely messages, using content-localization approach \cite{alzamzami2023osn,alzamzami2023content}. Content localization goes beyond translation which converts messages from one language to another. Content localization takes into account that context and tone of contents are transferred to the language and even to the dialect of interest; hence, we can exploit existing resources to solve the problem in hand without building the needed resources from scratch. This can be done through localizing the contents of an existing data resource (i.e. collected, cleaned, filtered, and annotated dataset) to a low-resource target language/dialect and using the localized content to train and build an OSB model in the language and/or dialect of interest.

\subsection{Spanish-French to Arabic Levantine-Gulf (SF-ArLG) Dataset}
Tweets in native Spanish and native French have been translated into Levantine and Gulf Arabic dialects. We require that all participants are professional translators, native in the Arabic dialects, and native or fluent in Spanish and French. Also, the translators are required to have a social involvement on social media (i.e. Twitter, Facebook, YouTube or any other platform). It is important to mention that the native Spanish and French tweets were originally translated from English tweets in DFMSD dataset \cite{dfsmd}.

For our translation, we adopt the content localization translation approach where translating the texts does not only convey a near-equivalent meaning but also addresses and integrates linguistic, cultural, tone, and contextual components of the texts. Same words might convey different meaning in different dialects; for example, the word “chips” refers to fried thin potato chips in North American English while it refers to “fries” in UK English.
The same applies for different dialects in Arabic language. An example of this, the word
"{\small \<صاحبي>}" in Gulf dialect refers to a friend while in Lebanese (i.e. Levantine) dialect it refers to boyfriend.
Since the dataset has been constructed based on social media (i.e. Twitter in this work) and
the tweets are mostly expressed in a day-to-day spoken language, the translations are customized to OSN cultural language. Thus, we take into consideration a number of additional criteria (i.e. in addition to the content localization based translation approach): (1) Consideration of OSN cultural language and expressions: iconic emotion (e.g. emoticons, emojis), slang abbreviations, and hashtag words. Hashtag
words are translated into the corresponding Arabic dialect while preserving their occurrence order and their context, (2) Consideration of Informal Language: informal language is used in casual settings like in daily-informal conversations; this includes slang words like "lol", "Oh mon dieu", or "bouffer", (3) Consideration of language code borrowing: code borrowing refers to using one primary language but mixing in words from another language to fit the primary language. For example, the word "lol" is used and written using the Arabic alphabet as "{\small \<لول>}".

The qualified Translators are provided with a subset of sentences and translation guidelines that cover the above criteria. In addition, they are advised to convert
mainly proper nouns into Arabic letters where applicable; for example names of people ("John" to "{\small \<جون>}"), and names of places ("UK" to "{\small \<بريطانيا>}"). The translators are also
advised to pay attention to the spelling as any misspelling would harm the quality of the
translation. Upon the completion of the translation task, the translators are asked to do a
round of proofreading before they submit the final translations. Note that the translation
is done for each dialect individually with the corresponding dialect translators. We have a total of $\approx$ 36,000 localized pairs (12,000 for each of French-Levantine, Spanish-Levantine, and Spanish-Gulf). The choice to study the two Arabic dialects is due to availability of some public native Levantine and Gulf OSB datasets like sentiment and hate speech that can be used to evaluate our proposed approach. In addition, the choice of three language-dialect pair is to meet the scope of this study
while respecting the writing space constraint of the paper.

\begin{figure}[h!]
\centering
\begin{multicols}{2}
\includegraphics[width=1.8\linewidth]{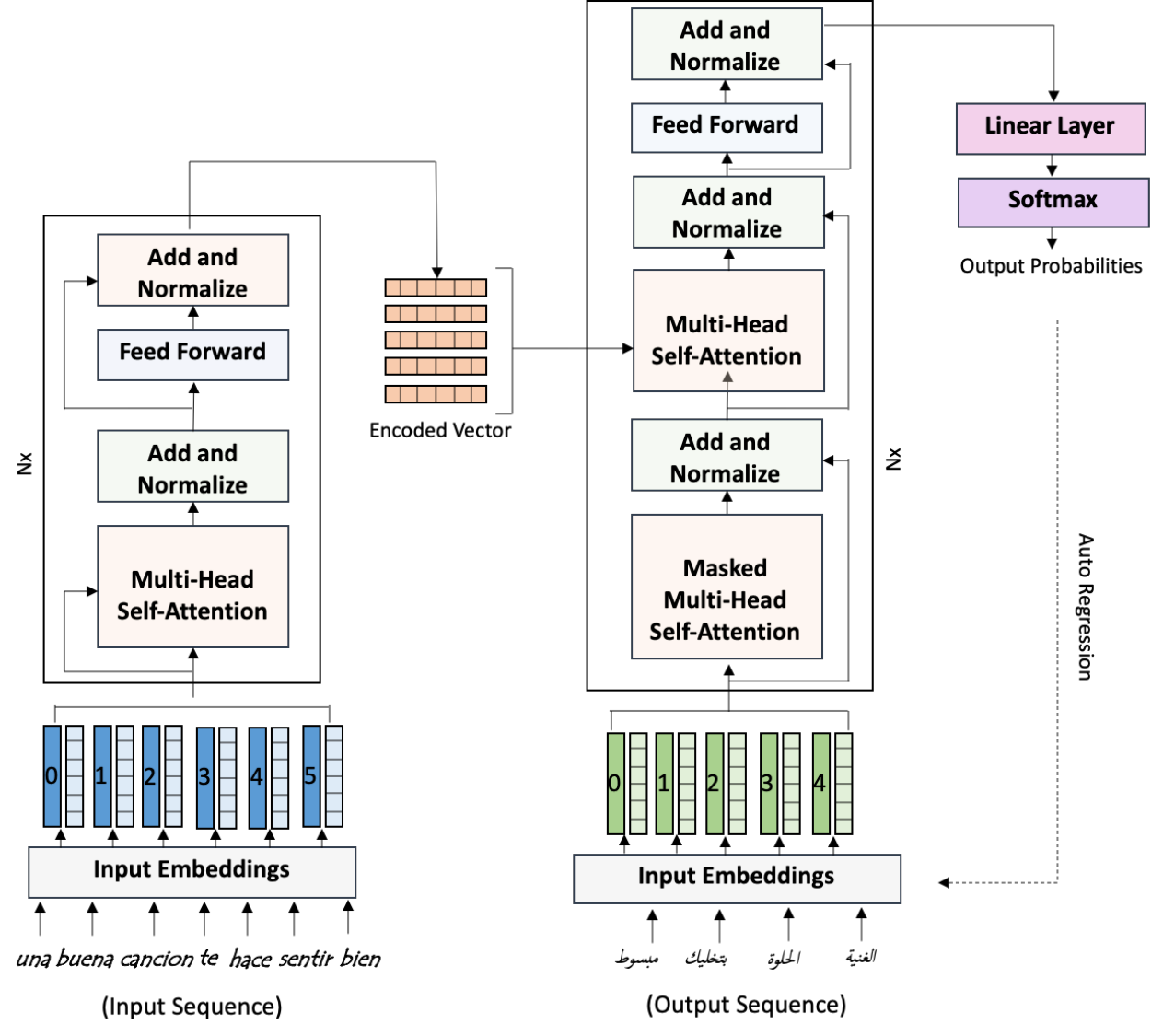}
\end{multicols}
\caption{Sequence-to-sequence Transformers architecture \cite{transformers} used in training our multi-lingual-dialectal NMT models.}
\label{fig:transformers-arch}
\end{figure}

\subsection{Transformers based Neural Machine Translation}

\begin{figure*}[h!]
\centering
\includegraphics[width=.7\textwidth]{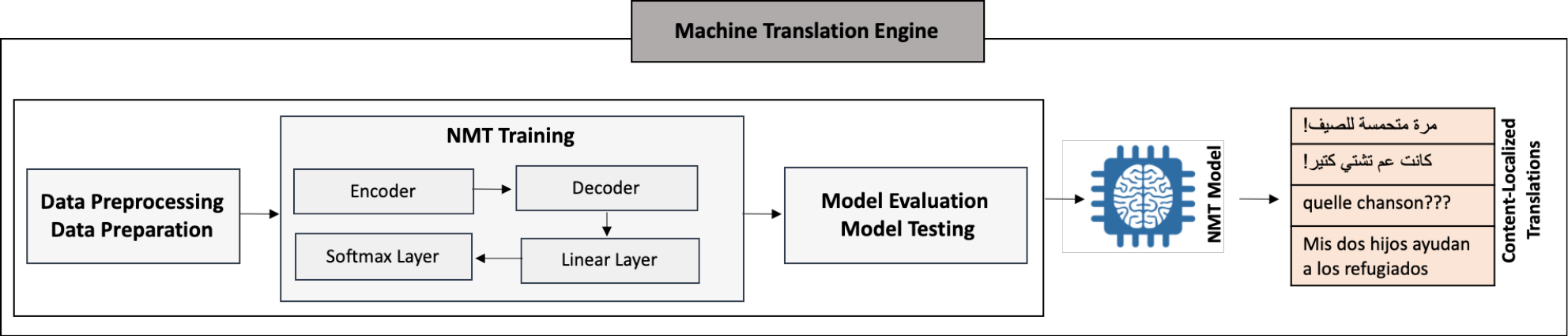}
\caption{Proposed methodology for neural machine translation.}
\label{fig:framework-nmt}
\end{figure*}

Figure. \ref{fig:framework-nmt} illustrates the methodology we follow in our neural machine translation modeling \cite{alzamzami2023osn}. Data is fed into the data preprocessing and preparation component; a set of filters are applied to clean unnecessary noise. The clean data is then fed into the neural machine translation (NMT) training component to start learning the translation between source language texts and their translation in the target language/dialect. Finally, the learnt model is evaluated and tested till it yields the best performance before it is ready to be used to translate new unseen data.

We use the sequence-to-sequence Transformers architecture depicted in Figure. \ref{fig:transformers-arch} \cite{liu2020multilingual}. It consists of 12 layers of encoder and 12 layers of decoders with model dimension of 1024 on 16 heads. On top of both encoder and decoder, there is an additional normalization layer that was found to stabilize the training. We use pretrained weights from mBART \cite{tang2020multilingual, liu2020multilingual} model that was pre-trained on 25 languages. It has been proven that transfer learning offers a rich set of benefits including improving the efficiency of model training and saving of time and resources since building a high-performance model from a scratch requires a large amount of data, time, resources, and efforts \cite{alzamzami2021monitoring, alzamzami2023transformer}. Therefore, we use the fine-tuning learning approach to train Transformers machine translation models as an attempt to solve the limitations of small datasets. For our Arabic dialects models, we use our proposed dataset to finetune the mBART \cite{liu2020multilingual} pretrained model. The model learns parallel translation from informal and slang French and Spanish to informal and dialectal Arabic. We have three models corresponding to two Arabic dialects: French-Levantine, Spanish-Levantine, and Spanish-Gulf.

\subsection{Online Social Behavior (OSB) Modeling}

\begin{figure*}[t!]
\centering
\includegraphics[width=.7\textwidth]{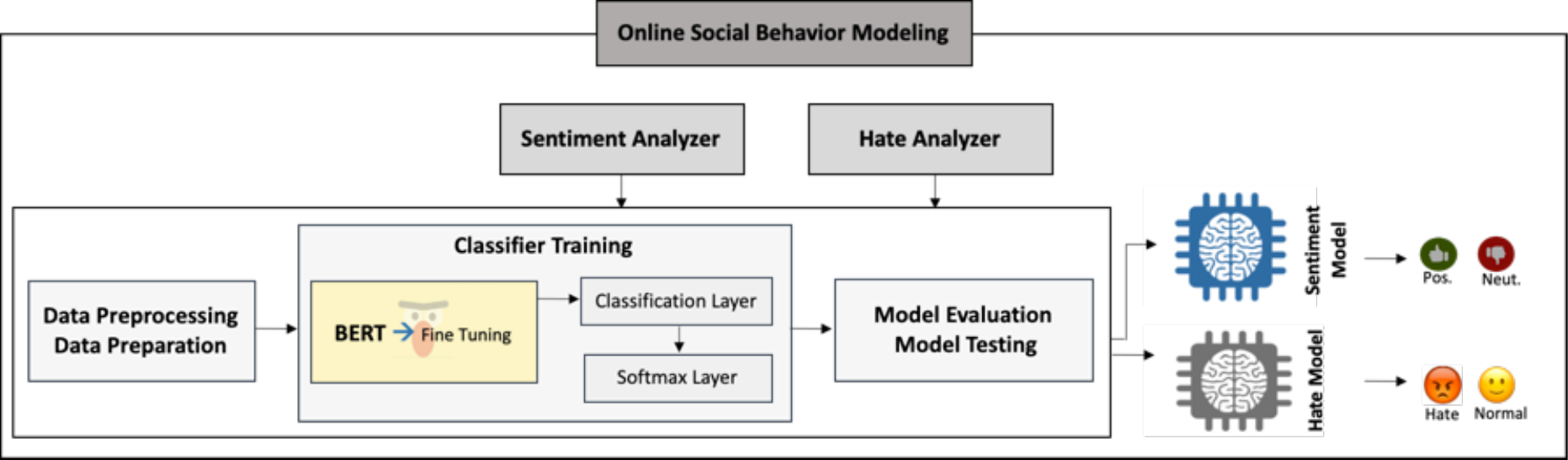}
\caption{Proposed methodology for online social behavior modeling.}
\label{fig:framework-osb}
\end{figure*}

Our study focuses on two types of online social behavior: sentiment and hate behaviors. Figure. \ref{fig:framework-osb} illustrates the proposed methodology followed in order to build sentiment and hate analyzers. Supervised classification approach is adopted in the modeling of OSB. The data preprocessing is performed first before the preprocessed data is fed into the training component. The Classifier Training component demonstrates the proposed neural network architecture. The BERT layer consists of BERT pre-trained embeddings which are representations of words and their relation to each other in n-dimension. BERT pre-trained model is fine-tuned by training the entire BERT architecture on our localized datasets in order to alleviate possible biases resulted from pre-training on Wikipedia corpus \cite{rietzler2019adapt}. BERT-base-arabic-camelbert-mix model \cite{inoue-etal-2021-interplay1} is used in this work; 
It consists of twelve layers and uses 110M parameters. The optimizer used is Adam with a learning rate of 1e-4, a weight decay of 0.01, learning rate warmup for 10,000 steps and linear decay of the learning rate after. 
the model is pre-trained on a mix of Modern Standard Arabic (MSA), dialectal Arabic (DA), and classical Arabic (CA).
A feed-forward neural network layer -used as a classification layer- is appended to the BERT layer. This classification layer produces logits that indicate the likelihood of a message belonging to a class. Soft max layer is used to normalize the logits and computes the probability of classes. The training is conducted by back propagating the errors throughout our architecture and updating the weights of the pre-trained weights and the weights of the appended layer based on our datasets. Early Stopping Approach is utilized to prevent over-fitting the neural network on the training data and improve the generalization of the models. Adams optimizer is used to optimize our models. Finally, we validate and test our models on the validation and test sets before generating the final result predictions. The prediction of sentiment analyzer is one of two polarity classes: positive or negative sentiments, whereas the hate analyzer predicts the existent or non-existent of hate content.

\section{Experiment Design and Evaluation Metrics}
\label{exp}

\subsection{Datasets and Preprocessing}
We list below the datasets we have used for modeling and evaluating our proposed approach: 
\begin{itemize}
    \item \textbf{French Twitter Sentiment Analysis Dataset (FTSAD) \cite{french-tweets-dataset}}:
We have randomly sampled 20,000 tweets with a balanced distribution between positive and negative classes. This dataset is used for modeling purposes.

    \item \textbf{ArSentD-Lev \cite{baly2019arsentd}}:
This dataset consists of 4,000 tweets (i.e. 1,232 positive tweets and 1,884 negative tweets) collected from the Arabic Levant region, and manually annotated through the crowd-sourcing approach. This dataset is used for evaluation purposes.
    \item \textbf{OCLAR datasets \cite{omari2022oclar}}:
OCLAR is an opinion corpus for Lebanese (i.e. Levantine) Arabic reviews.  

The positive class contains all reviews rating from 1 to 3 (3,465 reviews), while the negative class contains the reviews with rating values from 1 to 2 (451 reviews). This dataset is used for evaluation purposes.
    \item \textbf{Spanish HateSpeech Dataset \cite{wajidhassan-moosa-najiba_2022}}:
This dataset is a subset of a bigger multilingual hate speech dataset that consists of 13 languages, one of which is Spanish. We extracted the Spanish samples of total number of 12,423 texts, out of which 4,239 are labeled as hate speech. 

This dataset is used for modeling purposes.
    \item \textbf{L-HSAB datasets \cite{mulki2019hsab}}:
This dataset has been constructed and manually annotated for Levantine hate speech detection on social media. It contains a total of 5846 tweets, out of which 3,650 do not contain hate contents while 2,196 contain hate speech content. This dataset is used for evaluation purposes.
\end{itemize}

A list of preprocessing steps have been implemented to prepare the data before the modeling the stage.  For neural machine translation, we have applied: (1) Removing extra white-spaces, (2) Removing encoding symbols, (3) Removing URLs, (4) Converting text to lower case, (5) Removing tashkeel and harakat: tashkeel or harakat refer to all the diacritics placed over or below letters, (6) Normalizing Hamza.
For sentiment and hate speech modeling, the mentioned steps have been applied along with: (7) Removing user mention, (8) Removing special characters and numbers, (9) Removing stop-words.

\subsection{Experimental Design and Evaluation Protocol}
We design three scenarios to conduct the experiments of this paper as follows:\\
(1) We use our proposed (SF-ArLG) dataset to train three neural machine translation models: French-Levantine, Spanish-Levantine, and Spanish-Gulf. We randomly divided SF-ArLG into two groups: 90\% for training and 10\%for testing. Our training has been conducted through fine-tuning mBart \cite{tang2020multilingual} and using its default parameters.
For the evaluation metrics, we use BLEU and ROUGE metrics. BLEU metric is recommended for the translation tasks as it conducts robust assessment over the quality of translation fairly quickly. ROUGE complements BLEU in terms of evaluation where it focuses on recall; while BLEU is precision oriented. Therefore, we use F-score of BLEU and ROUGE.\\
(2) We localize an existing annotated sentiment dataset in French into a target language/dialect (i.e Arabic Levantine). Note that we preserve the source annotations of the source dataset. Then, we train a sentiment model using this localized dataset. Finally, we evaluate the trained model on an external dataset under the condition that the contents of the external dataset should be in the native target language/dialect. In this experiment, we translate the French sentiment dataset (FTSAD) \cite{french-tweets-dataset} to Arabic-Levantine using our proposed NMT French-Levant model. The localized dataset will be used later to train an Arabic-Levantine sentiment classifier. The purpose of this experiment is to examine the validity of our proposed approach that aims to minimize the dependency of language/dialects in modeling online social behavior (i.e. sentiment in this experiment).\\
(3) We localize an existing annotated hate speech dataset in Spanish into two target Arabic dialects, Levantine and Gulf, using our proposed NMT models. Note that we preserve the source annotations of the source dataset. Then, we train two hate classifiers using these localized datasets, one for each dialect. After that, we choose an external dataset of Arabic Levantine dialect. Finally, we evaluate the trained models on the external native Levantine dataset. The contents of this external dataset have been written and annotated in native Levantine dialect by native speakers. In this experiment, the Spanish hate dataset \cite{wajidhassan-moosa-najiba_2022} is localize to Arabic-Levantine and Arabic-Gulf dialects to be used later for training Arabic-Levantine and Arabic-Gulf hate classifiers. The purpose of this experiment is to investigate the effect of different dialects of the same language on the modeling and analysis of the online social behaviors on social media (i.e. hate speech in this experiment). 
    
For sentiment and hate speech modeling, the corresponding datasets have been randomly split into 80\% for training and 20\% for validation. We evaluate our models on the validation set during training- every 100 steps- to track its learning progress. We have implemented early-stopping approach to regularize the learning of the model during training in order to prevent any potential over-fitting to the training data. For evaluation measures, we have used accuracy, precision, recall, commonly used for classification evaluation, as evaluation metrics. Precision, recall  give a better view of model performance than accuracy alone does.
\section{Results and Analysis}
\label{res}

\subsection{Performance of the Proposed NMT Models}

Table. \ref{tab:fr-spanish-translators} presents the results of our Transformers-based NMT  models for French and Spanish to Arabic Levantine and Gulf dialects. Transformers models trained through fine-tuning a pretrained model (i.e. mBART in this work) are shown to yield a superior translation performance with at least F-score of 35 points.
Spanish to Arabic-Gulf NMT model is shown to have the highest learning performance (F-score of 37 points) followed by Spanish to Arabic-Gulf (F-scoe of 36 points) and then French to Arabic-Levantine (F-score of 35 points).  The nature of Transformers that use multi-head self-attention allows the models to learn word contextualization, and hence yield solid performance results in terms of F-score of BLUE and ROUGE. The self-attention enables contextualizing every word in various positions with respect to the whole sequence. This solves the homonym problem where similar words might have different meanings in different contexts. In addition, learning a model from scratch requires a huge size of data so that the model can converge properly. However, transfer learning technique has solved the problem related to limited and/or small data resources. As seen in Table. \ref{tab:fr-spanish-translators}, leveraging the transfer learning in learning our NMT models has tremendously yielded a high translation learning performance. 

\begin{table}
  \centering
  \caption{\small Performance results of three NMT models -in terms of F-score(BLEU, ROUGE)- using our propose dataset. The three models represents French to Arabic-Levantine, Spanish to Arabic-Levantine, and Spanish to Arabic-Gulf.}
    \resizebox{.4\textwidth}{!}{%

    \begin{tabular}{c|l|c}
    \toprule
    \multirow{4}[8]{*}{\begin{sideways}\textbf{Model}\end{sideways}} &       & \textbf{F-Score(BLEU, ROUGE)} \\
    \cmidrule{2-3}          & Fr-> Ar-Levantine & 35.2 \\
    \cmidrule{2-3}          & Es -> Ar-Levantine & 37.1 \\
    \cmidrule{2-3}          & Es -> Ar-Gulf & 36.2 \\
    \bottomrule
    \end{tabular}%
    }
  \label{tab:fr-spanish-translators}%
\end{table}%

 Figure. \ref{fig:fr-spanish_translations_examples} in the appendix, illustrates a sample of translations (i.e. generated by our proposed models) from/to French and Spanish to/from Arabic dialects (i.e. Levantine and Gulf in this study). It is important to note that all the sample translations were examined and approved by bilingual users with native to near-native bilingual language proficiency. The participant users approved that the contexts of the original messages were properly transferred -along with the corresponding dialect expression- into our generated translations. "Oh mon Dieu!" is an informal expression to say "oh my god or for god sake!" which was correctly translated into the Levantine expression "{\small \<ياربي!>}" that indicates an exclamation of anger, annoyance, or surprise. "bizarre", a common French word to describe someone or something as "weird or strange", was also correctly translated into a Levant term "{\small \<غريبة>}" meaning weird. The expression "{\small \<الفحص>}" is a very local way of saying "exam" in Levantine dialect, which could mean "check or examine" in other dialects like Gulf. Our Levantine model could successfully recognize it as "exam" and generate the sentence translation as "L'examen était très facile". "L'examen" in French refers to "exam" in English. Similarly, with expression "{\small \<تشتي>}" which means "raining" in the informal Levantine dialect; our model correctly translated it into French as "il pleuve tellement". "pleuve" is the French word for "raining".

\subsection{Performance of Sentiment Classification}
\begin{table}[htbp]
  \centering
  \caption{The performance of sentiment and hate models on the validation set in terms of accuracy, precision, and recall. The models represent French to Arabic-Levantine, Spanish to Arabic-Levantine, and Spanish to Arabic-Gulf.}
    \resizebox{.475\textwidth}{!}{%
    \begin{tabular}{c|c|p{7.165em}|c|c|c|c}
    \toprule
          & \multicolumn{1}{c}{} & \multicolumn{1}{c|}{} & \multicolumn{1}{p{4.835em}|}{\textbf{Validation \newline{}Precision}} & \multicolumn{1}{p{4.665em}|}{\textbf{Validation Recall}} & \multicolumn{1}{p{4.165em}|}{\textbf{Validation F-Score}} & \multicolumn{1}{p{4.415em}}{\textbf{Accuracy}} \\
\cmidrule{2-7}    \multicolumn{1}{c|}{\multirow{3}[6]{*}{\begin{sideways}\textbf{Model}\end{sideways}}} & \multicolumn{1}{p{4.415em}|}{\textbf{Sentiment}} & Fr->Ar-Gulf & 0.75  & 0.75  & 0.75  & 75 \\
\cmidrule{2-7}          & \multicolumn{1}{c|}{\multirow{2}[4]{*}{\textbf{Hate}}} & Es->Ar-Levantine & 0.69  & 0.68  & 0.68  & 69 \\
\cmidrule{3-7}          &       & Es->Ar-Gulf & 0.69  & 0.7   & 0.7   & 71 \\
    \bottomrule
    \end{tabular}%
    }
  \label{tab:mlmd-model-osb}%
\end{table}%

The results in Table. \ref{tab:mlmd-model-osb} present the performance of the proposed sentiment and hate models that were trained using the translated dataset (i.e. translated from Spanish and French to Arabic Levantine and Gulf dialects using our proposed NMT models). The results in Table. \ref{tab:mlmd-model-osb} depict the models' performances using the validation set split of the same data that the models were trained on.

\begin{figure}
\centering
\begin{subfigure}[b]{.23\textwidth}
  \centering
  \includegraphics[width=\linewidth]{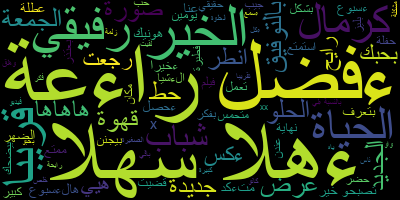}
  \caption{ Word-cloud for positive messages.}
  \label{fig:wc-sent-pos-fr-model1}
\end{subfigure} 
\begin{subfigure}[b]{.23\textwidth}
  \centering
  \includegraphics[width=\linewidth]{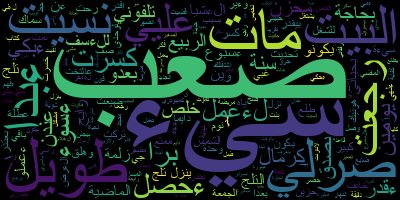}
  \caption{ Word-cloud for negative messages.}
  \label{fig:wc-sent-neg-fr-model1}
\end{subfigure}
\caption{Word-cloud generated from messages classified as positive or negative sentiment by our French to Arabic-Levantine sentiment model, using the translated French sentiment dataset \cite{french-tweets-dataset}.}
\label{fig:wc-sent-fr-model1}
\end{figure}

The French to Arabic-Levantine sentiment model is shown to have effectively learnt the sentiment classes (i.e. positive and negative) using the translated dataset (i.e. by our proposed NMT models). The sentiment model has scored 75\% of accuracy, 0.75 points of precision and recall. Figure. \ref{fig:wc-sent-fr-model1} illustrates the high frequency words used for positive and negative classes predicted by our sentiment model. The words in the figure corresponding to the positive class (Figure. \ref{fig:wc-sent-pos-fr-model1}))-, reflect positive sentiment such as "{\small \<رائعة>}" meaning "spectacular", "{\small \<الخير>}" meaning "good", "{\small \<رفيقي>}" meaning "my friend", "{\small \<أهلا وسهلا>}" meaning "greetings or welcome", "{\small \<هاهاها>}" meaning "hahaha", "{\small \<أفضل>}" meaning "better", "{\small \<الحلو>}" meaning "beautiful", "{\small \<بالتوفيق>}" meaning "good luck". Figure related to the negative class (Figure. \ref{fig:wc-sent-neg-fr-model1}) also reflect negative sentiment such as "{\small \<محزن>}" meaning "saddening", "{\small \<للأسف>}" meaning "unfortunately", "{\small \<أسوأ>}" meaning "worse", "{\small \<زعلان>}" meaning "sad or upset", "{\small \<صعب>}" meaning "difficult", "{\small \<غبي>}" meaning "stupid or dumb", "{\small \<أبكي>}" meaning "crying".  

While Table. \ref{tab:mlmd-model-osb} presents the performance of the models using the validation set, Table. \ref{tab:mlmd-osb-sent-external-ds} shows the performance of the sentiment model using an external dataset. The Levantine datasets (ArSentD-Lev \cite{baly2019arsentd} and OCLAR \cite{omari2022oclar}) are used to evaluate our Levantine sentiment model. The results show that the model is able to distinguish between positive and negative classes in both datasets; it has performed at a positive f-score of 0.81, negative f-score of 0.7, and over all accuracy of 77\%.

\begin{table}[h!]
  \centering
  \caption{\small The performance of French to Arabic-Levantine sentiment model, on external sentiment dataset, in terms of accuracy, precision, and recall.}
  \resizebox{.48\textwidth}{!}{%
    \begin{tabular}{c|p{13.415em}|c|c|c}
    \toprule
    \multicolumn{1}{r}{} & \multicolumn{1}{r|}{} & \textbf{F-Score (Positive)} & \textbf{F-Score (Negative)} & \textbf{Accuracy} \\
    \midrule
    \multirow{4}[8]{*}{\begin{sideways}\textbf{Model}\end{sideways}}   & Fr->Ar-Levantine\newline{} evaluated on \newline{}ArSentD-Lev \cite{baly2019arsentd} + OCLAR datasets \cite{omari2022oclar} & 0.81  & 0.7   & 77 \\
    \bottomrule
    \end{tabular}%
    }
  \label{tab:mlmd-osb-sent-external-ds}%
\end{table}%

\subsection{Performance of  Hate Speech Classification}

\begin{table}[htbp]
  \centering
  \caption{\small The performance of Spanish to Arabic-Levantine and Spanish to Arabic-Gulf hate models on an external hate speech dataset in Levantine dialect, in terms of accuracy, precision, and recall.}
  \resizebox{.48\textwidth}{!}{%
    \begin{tabular}{c|p{3.915em}|c|c|c|c|c|c|c}
    \toprule
    \multirow{5}[10]{*}{\begin{sideways}\textbf{Model}\end{sideways}} & \multicolumn{8}{c}{\textbf{Es --> Ar-Levantine evaluated on L-HSAB dataset \cite{mulki2019hsab}}} \\
\cmidrule{2-9}          & \multicolumn{1}{c|}{} & \multicolumn{3}{c|}{\textbf{Hate}} & \multicolumn{3}{c|}{\textbf{No-Hate}} &  \\
\cmidrule{2-9}          & \multicolumn{1}{c|}{} & \textbf{Precision} & \textbf{Recall} & \textbf{F-Score} & \textbf{Precision} & \textbf{Recall} & \textbf{F-Score} & \textbf{Accuracy} \\
\cmidrule{2-9}          & Es->Ar-Levantine & 0.66  & 0.68  & 0.67  & 0.8   & 0.79  & 0.8   & 75 \\
\cmidrule{2-9}          &  Es->Ar-Gulf & 0.7   & 0.54  & 0.61  & 0.76  & 0.86  & 0.8   & 74 \\
    \bottomrule
    \end{tabular}%
    }
  \label{tab:mlmd-hate-osb-external-es}%
\end{table}%

As seen in Table. \ref{tab:mlmd-model-osb}, our hate classifiers (i.e. Spanish-Arabic hate classifiers) have yielded solid performances using validation set split, in terms of accuracy
between 69\%-71\%. The overall precision and recall for both hate and non-hate classes are
quite high (i.e. at least 0.68 scores), which indicates that the classifiers have efficiently
learnt representative features that are able to detect the hate content correctly from the
data. We take a step further to assess our hate classifiers on real dataset that has been constructed and annotated manually in a native language and dialect; the Levantine L-HSAB dataset \cite{mulki2019hsab}, whose content is native Arabic Levant and has been collected from Levant geo-regions. The Spanish-Gulf model has shown a lower capability of detecting hate contents expressed in Levant dialect than its own dialect (i.e. Gulf). The Spanish-Gulf classifier has been able to recall only 54\% of hate messages compared to the Spanish-Levantine classifier that has recalled 67\% of hate contents at a precision score of 66\% while simultaneously keeping a high performance in distinguishing non-hate messages with as high as 80\% of f-score of precision and recall. Figure. \ref{fig:hate-tweet-sample} in the appendix, lists example hate-sentences expressed in Arabic-Levantine dialect, which the Spanish-Gulf classifier has classified as non-hate contents, while our Spanish-Levantine classifier has been able to correctly classify as hate contents.
As seen in the examples listed in the table, the same language has got different localized dialects; an expression in a certain dialect might mean something else in another and it might be used in a different context. Ignoring such a feature can negatively impact the learning models so much that they end up generating misleading outputs. For instance, the Levantine idiom "{\small \<كول هوا>}" -which means "eat dirt"- is a very local expression that Levantine people use in a negative situation (i.e. usually when in anger or disagreeing and it is used for swearing). The same applies to the toxic expressions "{\small \<مجرور, صباط, وعساك ما تطيب, هبيلة, ابو شحاطة, ازعر>}" - literally translated as "sewage, shoes, wish you stay unwell, dumb, father of thong-sandal, troubling person"- are used exclusively by Levantine people to express anger or dissatisfaction.
\section{Conclusion}
\label{conc}
This work proposes a framework to localize contents of messages between high and low resource languages/dialects. The proposed framework ensures transferring the context from a source language to target language/dialects. We propose the first of its kind -to the best of our knowledge- an informal parallel translation dataset (SF-ArLG dataset) from/to Spanish and French to/from Arabic Levantine and Gulf dialects; it is an attempt to minimize the dependency of languages in modeling online social behaviors on social media. By using the localized resources of high-resource languages, we have been able to build reliable sentiment and hate speech classifiers for low-resource language/dialects (i.e. Arabic Leavantine and Gulf dialects in this study) without building new Levant and Gulf resources from scratch.
The results of our comprehensive experiments confirm the quality of our proposed dataset and that is shown in the NMT models' ability to learn the translation of the dialectal Arabic. Moreover, our localized sentiment and hate models are shown to be able to effectively learn sentiment and hate speech from our localized data, and to successfully distinguish between positive and negative classes as well as detecting hate content from data which, in turn, proves the effectiveness of our NMT models that have shown solid capability of transferring the context of informal-social messages from a language to another language/dialects. Also, our finding highlights the importance of distinguishing dialects of the same language and their localized contextual meanings. Overlooking those differences results in inaccurate understanding of the target dialect, which in turn leads to misleading and imprecise analysis of online social behaviors as seen in our Levant hate model being able to detect hate content from Levant messages while the Gulf hate model has struggled to do so. The results and findings of this study have shown the potential of the proposed approach to expand the range of online social behavior analysis in languages/dialects that lack proper data resources for online social behaviors. 

\newpage

\appendix
\section*{Appendix}
\label{sec:appendix}

\begin{figure}[h!]
\centering
\includegraphics[width=.99\linewidth]{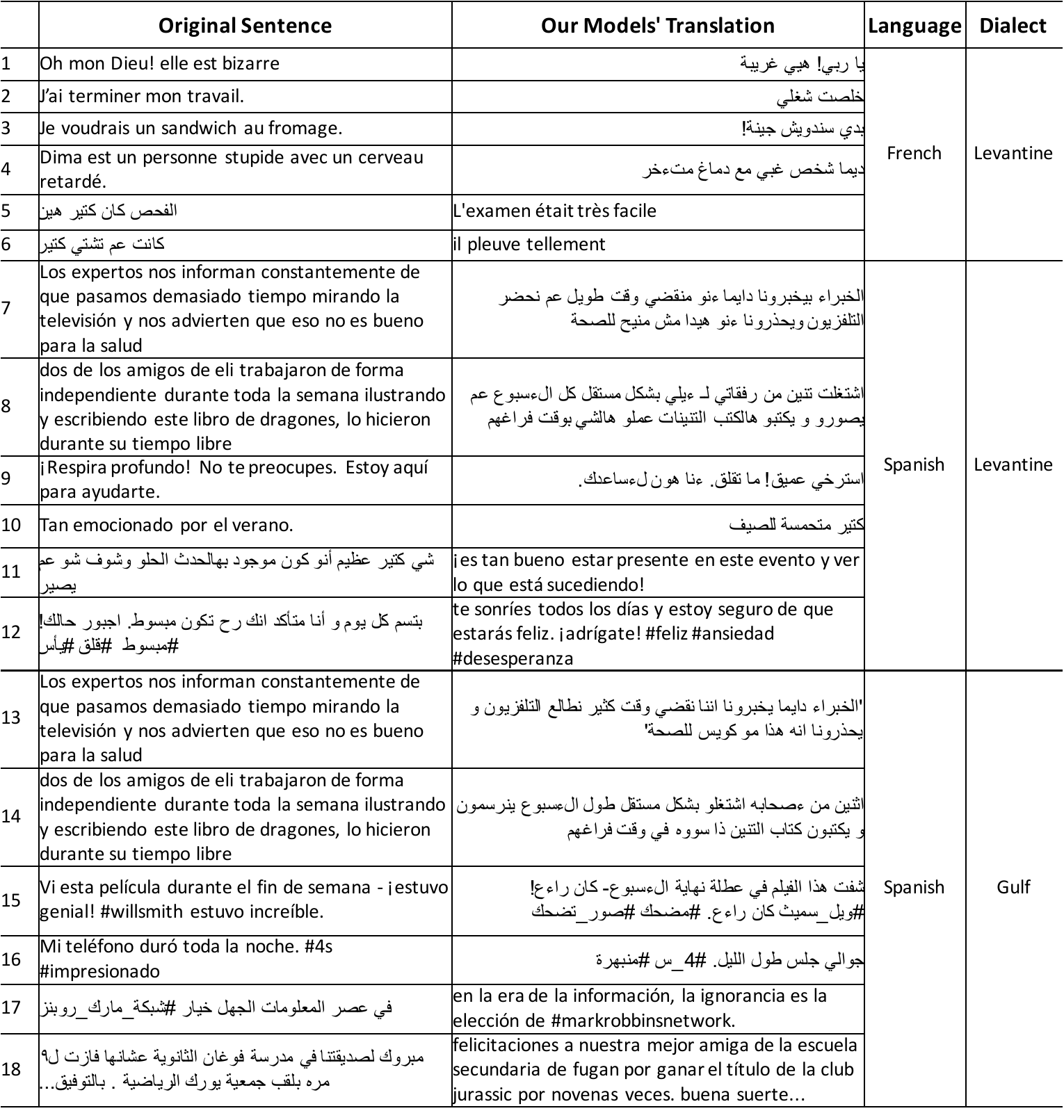}
\caption{A sample of generated translations by our proposed NMT models using the transitive translation approach. The translations are from three models: French to/from Arabic-Levantine, Spanish to/from Arabic-Levantine, and Spanish to/from Arabic-Gulf.}
\label{fig:fr-spanish_translations_examples}
\end{figure}

\newpage

\begin{figure}[h!]
\centering
\includegraphics[width=1\linewidth]{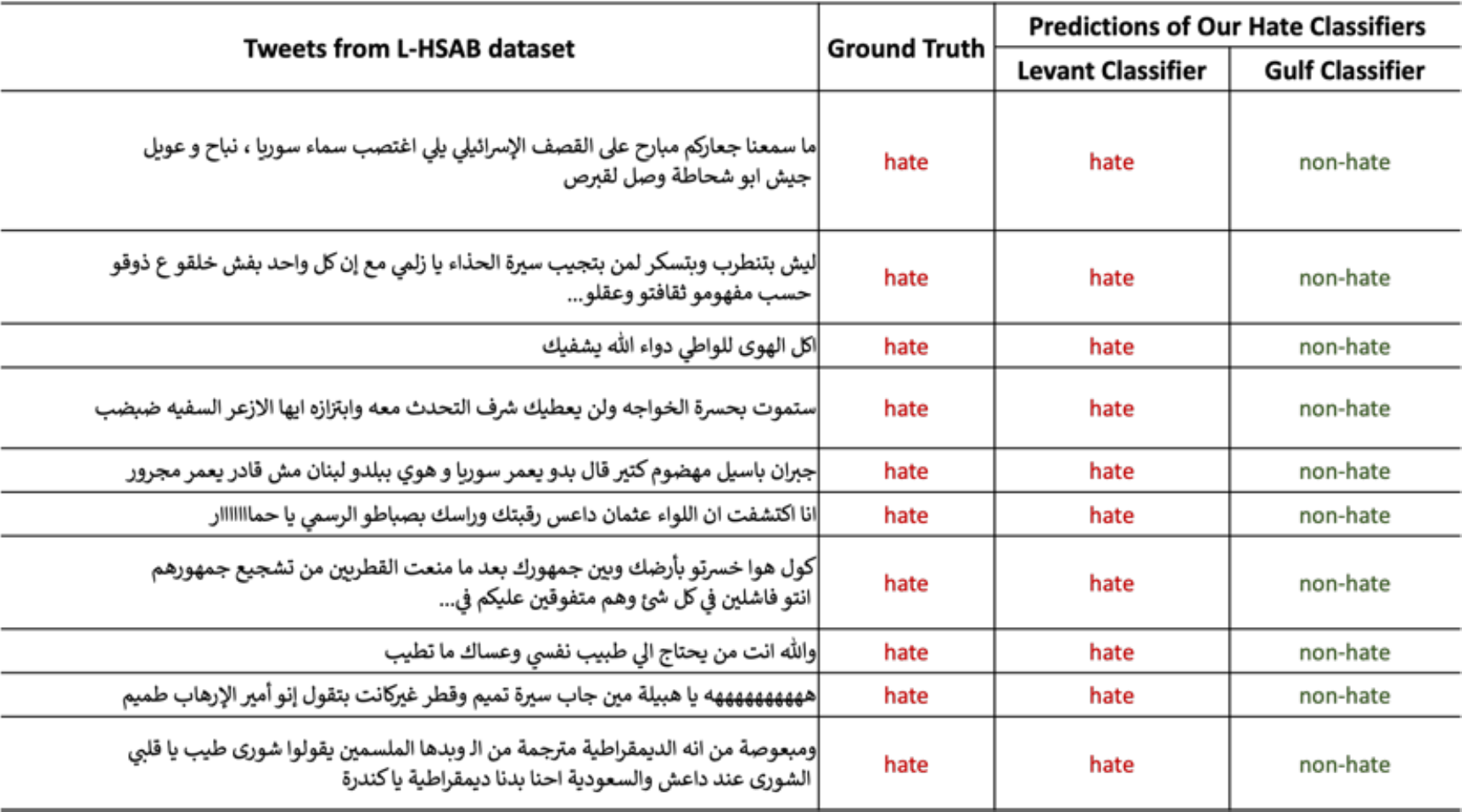}
\caption{A sample of tweets correctly classified as hate speech by our Levant hate classifier and incorrectly classified and non-hate speech by Gulf hate classifier.}
\label{fig:hate-tweet-sample}
\end{figure}

\bibliographystyle{unsrt}  
\bibliography{references}  

\end{document}